\documentclass[10pt,letterpaper]{article}
\usepackage[top=0.85in,left=2.75in,footskip=0.75in]{geometry}

\usepackage{amsmath,amssymb}

\usepackage{changepage}
\usepackage{subfigure}
\usepackage[utf8x]{inputenc}
\usepackage{multirow}
\usepackage{textcomp,marvosym}

\usepackage{cite}

\usepackage{nameref,hyperref}

\usepackage[right]{lineno}

\usepackage{microtype}
\DisableLigatures[f]{encoding = *, family = * }

\usepackage[table]{xcolor}

\usepackage{array}

\newcolumntype{+}{!{\vrule width 2pt}}

\newlength\savedwidth

\raggedright
\usepackage[aboveskip=1pt,labelfont=bf,labelsep=period,justification=raggedright,singlelinecheck=off]{caption}

\makeatletter
\renewcommand{\@biblabel}[1]{\quad#1.}
\makeatother

\usepackage{lastpage,fancyhdr,graphicx}
\usepackage{epstopdf}
\fancyheadoffset[L]{2.25in}
\fancyfootoffset[L]{2.25in}
\begin{document}
\vspace*{0.2in}

\begin{flushleft}
{\Large
\textbf\newline{EMDS-5: Environmental Microorganism Image Dataset Fifth Version for 
Multiple Image Analysis Tasks} % Please use "sentence case" for title and headings (capitalize only the first word in a title (or heading), the first word in a subtitle (or subheading), and any proper nouns).
}
\newline
\\
Zihan Li\textsuperscript{1},
        Chen Li\textsuperscript{1},
        Yudong Yao\textsuperscript{2}, 
        Jinghua Zhang\textsuperscript{1},
        Md Mamunur Rahaman\textsuperscript{1}, 
        Hao Xu\textsuperscript{1},
        Frank Kulwa\textsuperscript{1},
        Bolin Lu\textsuperscript{3},
        Xuemin Zhu\textsuperscript{4},
        Tao Jiang\textsuperscript{5}
\\
\bigskip
\textbf{1} Microscopic Image and Medical Image Analysis Group, MBIE College, Northeastern University, 110169, Shenyang, PR China
\\
\textbf{2} Department of Electrical and Computer 
Engineering, Stevens Institute of Technology, Hoboken, NJ 07030, USA
\\
\textbf{3} School of Biomedical Engineering, Huazhong 
University of Science and Technology, Wuhan 430074, China
\\
\textbf{4} Whiting School of Engineering, Johns Hopkins 
University, 500 W University Parkway, MD, USA 21210, USA
\\
\textbf{5} School of Control Engineering, Chengdu 
University of Information Technology, Chengdu 610225, China
\\
\bigskip

Corresponding author: Chen Li, E-mail: lichen201096@hotmail.com

\end{flushleft}
\section*{Abstract}
\emph{Environmental Microorganism Data Set Fifth Version} (EMDS-5) is a microscopic image dataset including original \textit{Environmental Microorganism} (EM) images and two sets of \emph{Ground Truth} (GT) images. The GT image sets include a single-object GT image set and a multi-object GT image set. The EMDS-5 dataset has 21 types of EMs, each of which contains 20 original EM images, 20 single-object GT images and 20 multi-object GT images. EMDS-5 can realize to evaluate image preprocessing, image segmentation, feature extraction, image classification and image retrieval functions. In order to prove the effectiveness of EMDS-5, for each function, we select the most representative algorithms and price indicators for testing and evaluation. The image preprocessing functions contain two parts: image denoising and image edge detection. Image denoising uses nine kinds of filters to denoise 13 kinds of noises, respectively. In the aspect of edge detection, six edge detection operators are used to detect the edges of the images, and two evaluation indicators, peak-signal to noise ratio and 
mean structural similarity, are used for evaluation. Image segmentation includes single-object image segmentation and multi-object image segmentation. Six methods are used for single-object image segmentation, while $k$-means and U-net are used for multi-object segmentation.We extract nine features from the images in EMDS-5 and use the Support Vector Machine classifier for testing. In terms of image classification, we select the VGG16 feature to test different classifiers. We test two types of retrieval approaches: texture feature retrieval and deep learning feature retrieval. We select the last layer of features of these two deep learning networks as feature vectors. We use mean average precision as the evaluation index for retrieval.

\section{Introduction}
\label{sec:introduction}
\subsection{Environmental Microorganisms}
\begin{figure}[htbp!]
\centering
\subfigure[ \emph{Rotifera}]
{\label{fig:EMS:a}
\includegraphics[width=0.4\hsize]{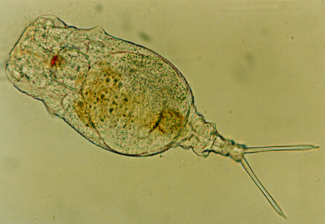}}
\subfigure[ \emph{Arcella}]
{\label{fig:EMs:b}
\includegraphics[width=0.41\hsize]{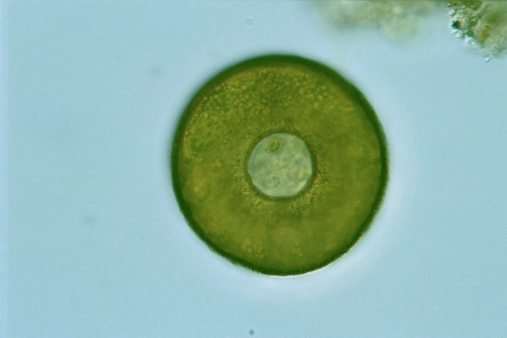}} 
\caption{An example of EM images.}
\label{fig:EMs} 
\end{figure}

All the time, \textit{Environmental Microorganisms} 
(EMs)~\cite{li-2016-environmental} are part of our environment. Some EMs 
bring us benefits, while others affect our physical health. Many researchers 
devote themselves to study these microorganisms to improve our lives. 
Nowadays we usually use a microscope to observe EMs. However, scholars 
sometimes get it wrongly. Image analysis has a great significance for the 
analysis of EM images. It can help researchers to analyze the types and 
forms of EMs. For example, \textit{Rotifera} is a common EM and it is widely 
distributed in lakes, ponds, rivers, rivers and other brackish water bodies, 
having great significance in the study of ecosystem structure function and 
biological productivity because of their extremely fast reproduction rate 
and high yield. In addition, \textit{Arcella} is also a kind of common EMs. 
\textit{Arcella} mainly feeds on plant giardia and single-celled algae. An 
oligoplastic water body is the most suitable living environment. Two EM image 
examples are shown in Fig.~\ref{fig:EMs}.
\subsection{Environmental Microorganisms Application Scenarios} 
The principal sources of noise in digital images arise during image 
acquisition or transmission~\cite{Gonzalez-2007-Digital}. Image denoising 
can reduce the noise of EM images while preserving the details. In addition, 
image segmentation is a process of partitioning the image into some 
non-intersecting regions such that each region is homogeneous and the union 
of no two adjacent regions is homogeneous~\cite{pal-1993-review}. So image 
segmentation technology can be used to segment images of EMs to separate 
microorganisms from the complex background of the images. After, the feature 
extraction part is performed. When the input data to an algorithm is too 
large to be processed and suspected to be redundant (much data but not much 
information), the input data will be transformed into a reduced set of 
features (such as usually used ``feature vectors''). Feature extraction is 
the process of converting input data into set of 
features~\cite{Guyon-2006-Feature}. For the segmented EMs, we usually extract 
their shape features, color features or deep learning features. We need to 
use these features for image classification and image retrieval. Image 
classification is determined by the trained classifier, which is trained by 
the training data with category labels. We put the extracted feature vectors 
into a classifier and match them with the known data and put them into the 
same group of EMs. Image retrieval is given a query image and searches for 
similar images. We extract the feature vector and calculate its similarity 
to the feature vector of the known data.

\subsection{Contribution}
Environmental investigations are always operated in outdoor environments, 
where conditions like temperature and salinity are changing continuously. 
Because EMs are very sensitive to these conditions, their quantity is 
easily influenced. It is difficult to collect sufficient EM 
images~\cite{Kosov-Environmental}. As a result, when researchers want to 
create EM datasets, they often run out of data. Currently, there are some 
existing EM datasets, but many of them are not open source. This will make 
it difficult for EM researchers to obtain the existing EM data set and 
require much time to collect it. \emph{Environmental Microorganism Data Set 
Fifth Version} (EMDS-5) will be made available to other researchers as an 
open source dataset. In addition, EMDS-5 has many advantages over other 
datasets. EMDS-5 will provide the corresponding \textit{Ground Truth} (GT). 
Since it takes a lot of time and human resources to make GT images, many 
datasets do not make GT images corresponding to their own data sets. 
GT images play an important role in image analysis. GT images can be a 
significant evaluation index for image segmentation. The result of image 
segmentation can be judged by comparing the segmented image with the GT images. 
EDMS-5 has a variety of EM images to provide sufficient data support for 
image classification and image retrieval. The experiment of 
multi-classification can be carried out for image classification of 
multi-species EMs to obtain ideal results. At the same time, many kinds of 
EM images and sufficient data provide strong data support for the results 
of image retrieval.

\subsection{Related work}
To the best of our knowledge, we know seven special EM datasets. In two cases, 
we only know the types of microorganisms and the number of samples used in 
user experiments. The remaining five are our EMDS series. The seven 
databases are NMCR, CECC, EMDS-1, EMDS-2, EMDS-3, EMDS-4 and EMDS-5. 
The information we obtain from the datasets is presented in 
TABLE~\ref{tab:EM-datasets}.
\begin{table}[htbp!]
\small
\centering
\caption{Basic information of known open source EM image databases. 
Dataset (DS), Number of classes (NoC), Number of EM images (NoEI), 
Number of GT images (NoGI), Single-object (SO), Multi-object (MO).}
\begin{tabular}{
| p{0.1\textwidth}<{\centering}
| p{0.1\textwidth}<{\centering}
| p{0.1\textwidth}<{\centering}
| p{0.1\textwidth}<{\centering}
| p{0.1\textwidth}<{\centering}
| p{0.1\textwidth}<{\centering}|}%
\hline
    DS & Time & NoC & NoEI & NoGI & Related work \\
\hline
      NMCR & 2007 & 8 & 640 & - & \cite{Li-2007-A}\\
\hline
      CECC & 2009 & 4 & 64 & - & \cite{Li-2009-An}\\
\hline      
      \multirow{3}{*}{EMDS-1} &\multirow{3}{*}{2013} &\multirow{3}{*}{10} &\multirow{3}{*}{200} & \multirow{3}{*}{SO: 200} & \cite{Li-2013-Classification}\cite{Li-2013-A} \cite{Li-2016-Content-based}\\
\hline
      \multirow{3}{*}{EMDS-2} & \multirow{3}{*}{2013} & \multirow{3}{*}{10} & \multirow{3}{*}{200} & \multirow{3}{*}{SO: 200} &\cite{Li-2013-Classification} \cite{Li-2013-A} \cite{Li-2016-Content-based}\cite{yang-2014-shape} \cite{Li-2015-Application}\\
\hline
      \multirow{3}{*}{EMDS-3} & \multirow{3}{*}{2015} & \multirow{3}{*}{15} & \multirow{3}{*}{300} & \multirow{3}{*}{SO: 300} &\cite{li-2015-environmental}\cite{li-2016-environmental} \cite{li-2017-content}\\
\hline
      \multirow{2}{*}{EMDS-4} & \multirow{2}{*}{2015} & \multirow{2}{*}{21} & \multirow{2}{*}{420} &\multirow{2}{*}{SO: 420} &\cite{zou-2016-environmental}\cite{Kosov-2018-Environmental} \cite{zou_2016_environmental}\cite{zou-2016-content}\\
\hline
      \multirow{2}{*}{EMDS-5} & \multirow{2}{*}{2019} & \multirow{2}{*}{21} & \multirow{2}{*}{420} & SO:420, MO:420&\cite{Zhang-2020-A} \cite{li-mrfu}\\
\hline
\end{tabular}
\label{tab:EM-datasets}
\end{table}

\section{Dataset Information of EMDS-5 }
\begin{figure*}[htbp!]
\centering
\subfigure{
\includegraphics[width=0.31\hsize]{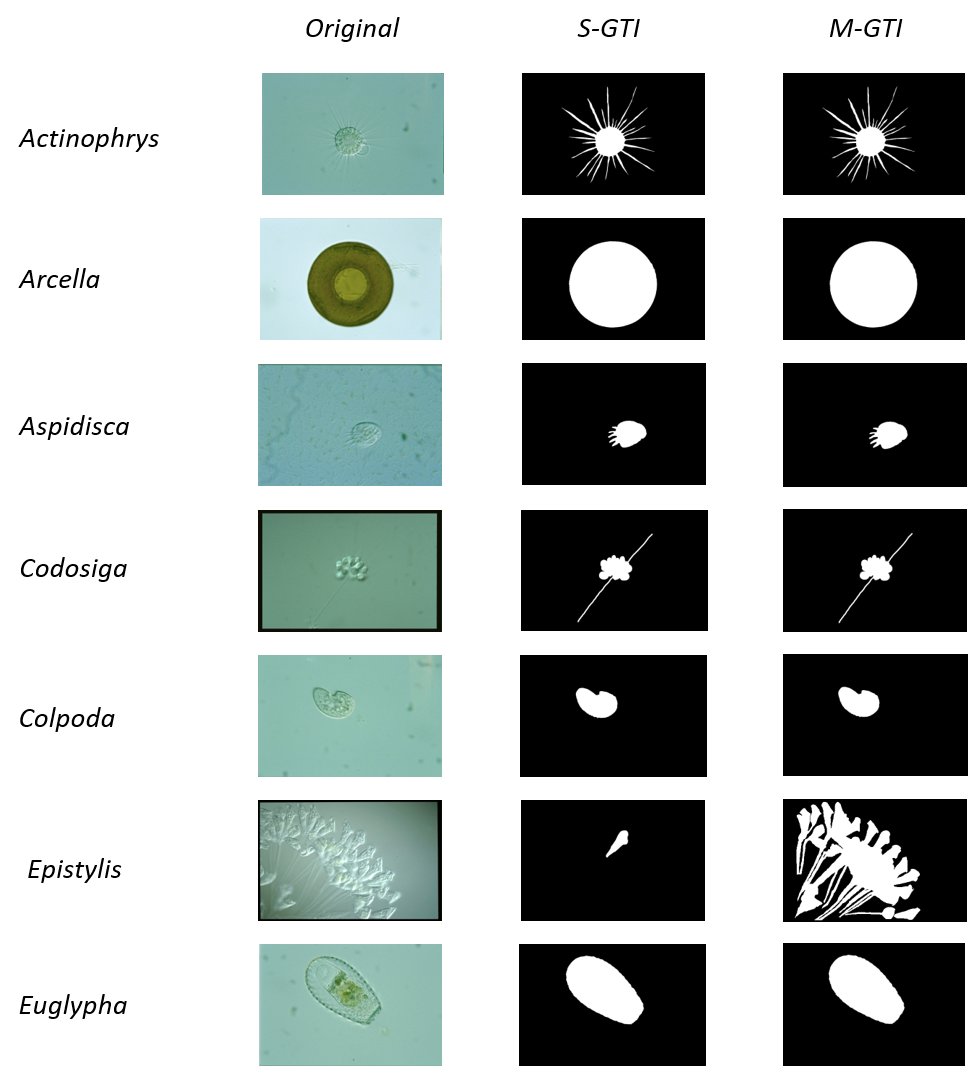}
}%
\subfigure{
\includegraphics[width=0.31\hsize]{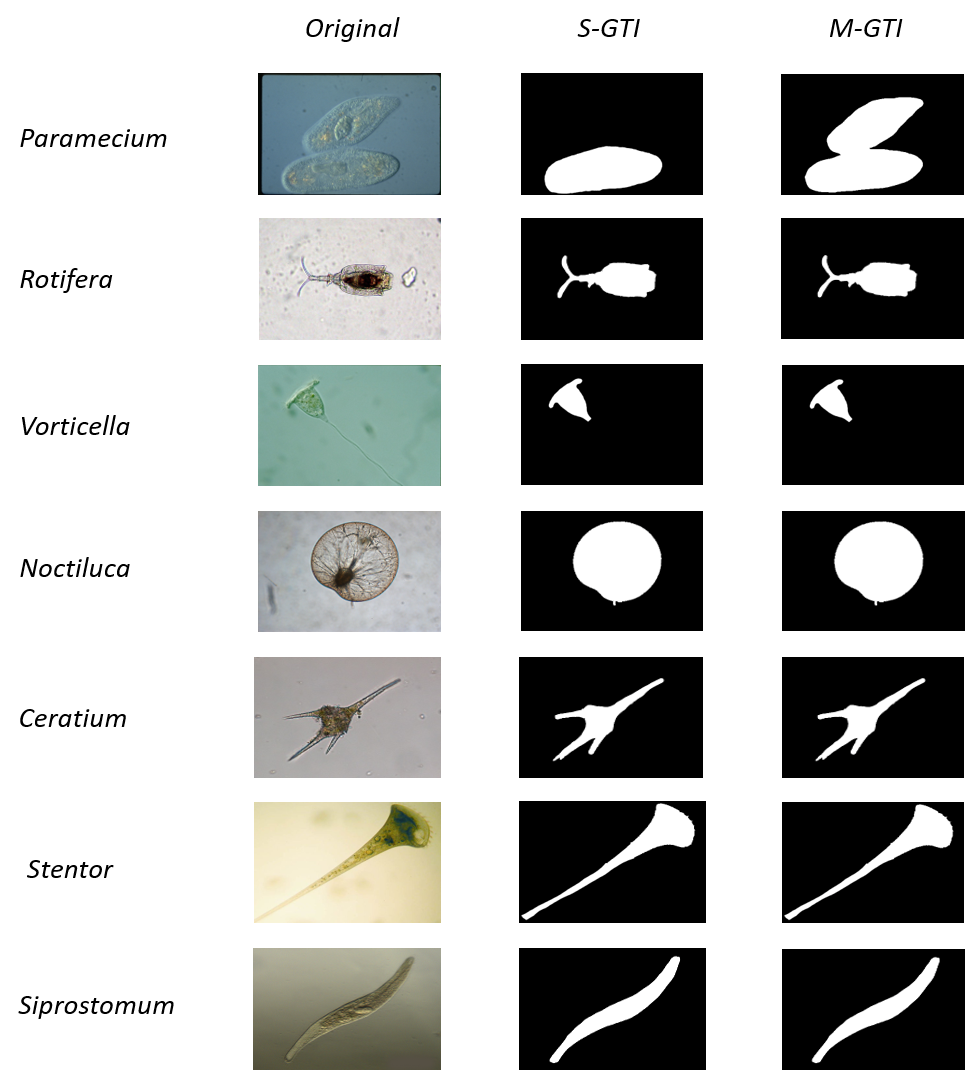}
}%
\subfigure{
\includegraphics[width=0.31\hsize]{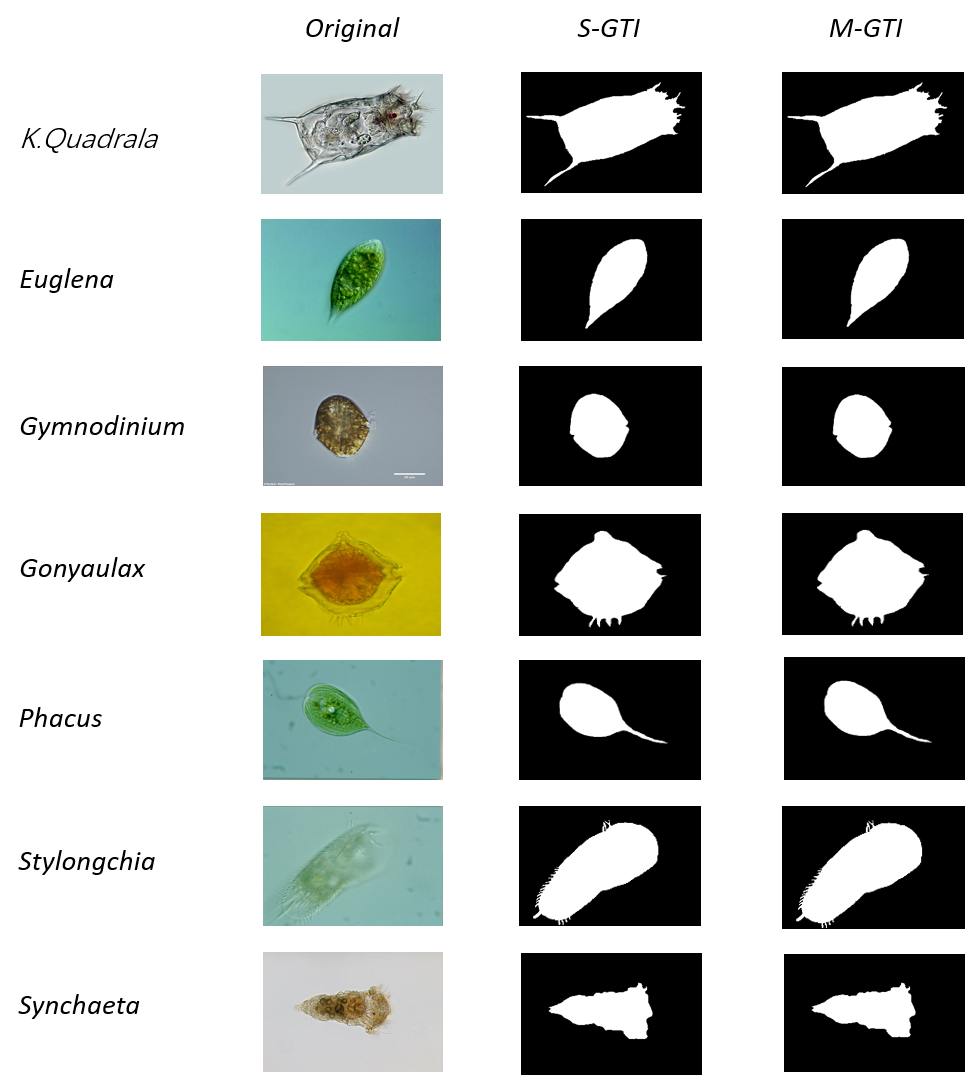}
}%
\centering
\caption{An example of 21 EM classes in EMDS-5. 
Single-object GT images (SGI), Multi-object GT images (MGI).}
\label{fig:EMDS5-examples}
\end{figure*}

EMDS-5 is made up of 1260 images of 21 EM classes. The original 420 EM images 
are partly collected under artificial light sources and partly under natural 
light sources with a $400 \times$ optical microscope. In addition, 840 GT 
images are manually prepared, including 420 single-object GT images and 
420 multi-object GT images. Basic information of the 21 EM classes in EMDS-5 
is given in TABLE~\ref{tab:list}, and an example of 21 EM classes in EMDS-5 
is shown in Fig.~\ref{fig:EMDS5-examples}.
\begin{table}[htbp!]
\scriptsize
\centering
\caption{Basic information of 21 EM classes in EMDS-5. 
Number of original images (NoOI), Number of single-object GT images (NoSGI), 
Number of multi-object GT images (NoMGI), Visible characteristics (VC).}
\begin{tabular}{
| p{0.1\textwidth}<{\centering}
| p{0.03\textwidth}<{\centering}
| p{0.05\textwidth}<{\centering}
| p{0.05\textwidth}<{\centering}
| p{0.09\textwidth}<{\centering}
| p{0.12\textwidth}<{\centering}
| p{0.03\textwidth}<{\centering}
| p{0.05\textwidth}<{\centering}
| p{0.05\textwidth}<{\centering}
| p{0.1\textwidth}<{\centering}|}
\hline
      Classes & NoOI & NoSGI &  NoMGI & VC & Classes & NoOI & NoSGI &  NoMGI & VC\\
\hline
      \multirow{3}{*}{\textit{Actinophrys}} & \multirow{3}{*}{20} & \multirow{3}{*}{20} & \multirow{3}{*}{20} & \multirow{3}{*}{Spherical} &\multirow{3}{*}{\textit{Ceratium}} & \multirow{3}{*}{20} & \multirow{3}{*}{20} & \multirow{3}{*}{20} & Ring or slightly spiral\\
      \hline
      \textit{Arcella} & 20 & 20 & 20 & Ellipsoid & \textit{Stentor} & 20 & 20 & 20 & Trumpet\\
      \hline
     \multirow{4}{*}{\textit{Aspidisca}} & \multirow{4}{*}{20} & \multirow{4}{*}{20} & \multirow{4}{*}{20} & Parasiticon skin lesions of Turbot & \multirow{4}{*}{\textit{Siprostomum}} & \multirow{4}{*}{20} & \multirow{4}{*}{20} & \multirow{4}{*}{20} & \multirow{4}{*}{Worm-like}\\
      \hline
      \multirow{3}{*}{\textit{Codosiga}} & \multirow{3}{*}{20} & \multirow{3}{*}{20} & \multirow{3}{*}{20} & Each cell has a flagella & \multirow{3}{*}{\textit{K.Quadrala}} & \multirow{3}{*}{20} & \multirow{3}{*}{20} & \multirow{3}{*}{20} & \multirow{3}{*}{Ellipsoid}\\
      \hline
      \textit{Colpoda} & 20 & 20 & 20 & Kidney & \textit{Euglena} & 20 & 20 & 20 & Phototaxis\\
      \hline
      \multirow{2}{*}{\textit{Epistylis}} & \multirow{2}{*}{20} & \multirow{2}{*}{20} & \multirow{2}{*}{20} & Funnel shape & \multirow{2}{*}{\textit{Gymnodinium}} & \multirow{2}{*}{20} & \multirow{2}{*}{20} & \multirow{2}{*}{20} & Spherical or oval\\
      \hline
      \multirow{5}{*}{\textit{Euglypha}} & \multirow{5}{*}{20} & \multirow{5}{*}{20} & \multirow{5}{*}{20} & \multirow{5}{*}{Oval} & \multirow{5}{*}{\textit{Gonyaulax}} & \multirow{5}{*}{20} & \multirow{5}{*}{20} & \multirow{5}{*}{20} & Covered by tightly bonded cellulosic plates\\
      \hline
      \multirow{3}{*}{\textit{Paramecium}} & \multirow{3}{*}{20} & \multirow{3}{*}{20} & \multirow{3}{*}{20} & \multirow{3}{*}{Sole shape} & \multirow{3}{*}{\textit{Phacus}} & \multirow{3}{*}{20} & \multirow{3}{*}{20} & \multirow{3}{*}{20} & Dorsal ventral flat\\
      \hline
      \multirow{3}{*}{\textit{Rotifera}} & \multirow{3}{*}{20} & \multirow{3}{*}{20} & \multirow{3}{*}{20} & Roulette composed of cilia &  \multirow{3}{*}{\textit{Stylonychia}} & \multirow{3}{*}{20} & \multirow{3}{*}{20} & \multirow{3}{*}{20} & \multirow{3}{*}{Fan shaped}\\
      \hline
     \multirow{3}{*}{ \textit{Vorticlla}} & \multirow{3}{*}{20} & \multirow{3}{*}{20} & \multirow{3}{*}{20} & \multirow{3}{*}{Dendritic} & \multirow{3}{*}{\textit{Synchaeta}} & \multirow{3}{*}{20} & \multirow{3}{*}{20} & \multirow{3}{*}{20} & Transparent and flexible\\
      \hline
      \multirow{3}{*}{\textit{Noctiluca}} & \multirow{3}{*}{20} & \multirow{3}{*}{20} & \multirow{3}{*}{20} & With luminous ability & -& -& -& -& - \\
\hline
      Total & 420 & 420 & 420 & - & Total & 420 & 420 & 420 & -\\
\hline
\end{tabular}
\label{tab:list}
\end{table}

Three researches from University of Science and Technology Beijing (China) and 
University of Heidelberg (Germany) provide the original image data of EMDS-5. 
Furthermore, the preparation of EMDS-5 GT images is jointly completed by 
three researchers from Northeastern University (China), Johns Hopkins 
University (US) and Huazhong University of Science and Technology (China). 
All of them have research backgrounds in Environmental Engineering or 
Biological Information Engineering. 
Especially, EMDS-5 GT images are manually labelled based on pixel-level 
based on following two rules: 
\begin{itemize}
\item Rule A: The area where an EM is located is labelled as foreground 
(1, white). 
In contrast, other areas are labelled as background (0, black). 

\item Rule B: Because the microscopic images in the EMDS-5 dataset are 
collected under optical microscopes, this process produces interference 
fringes and results in unwanted edges in the EM images. Hence, when making 
GT images, the most complicated thing is to determine the edges of an EM. 
First, each researcher selects the edges that she or he thinks are the 
clearest to label. Then, if their labelling results are conflict, they have 
a collective discussion to judge and decide a final solution.
\end{itemize} 

\section{Image Processing Evaluation Using EMDS-5}
\subsection{Evaluation of Image Denoising Methods}
\begin{figure}[htbp!]
\centering
\subfigure[Pepper]{
\includegraphics[width=0.3\hsize]{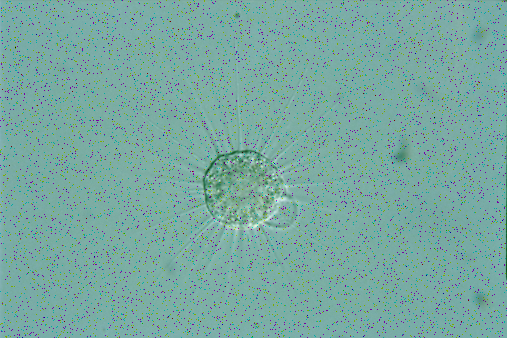}
}%
\quad
\subfigure[Poisson]{
\includegraphics[width=0.3\hsize]{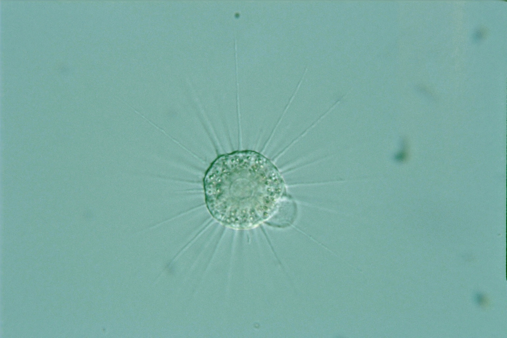}
}%
\quad               
\subfigure[Pepper and salt: 0.01 density ]{
\includegraphics[width=0.3\hsize]{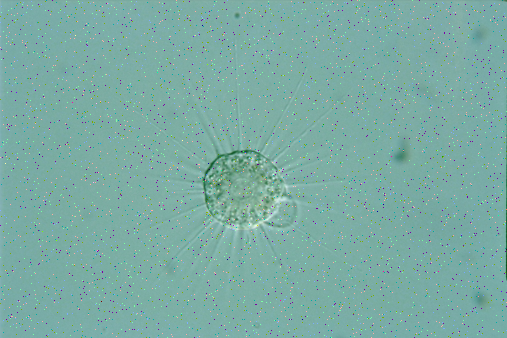}
}%
\quad 
\subfigure[pepper and salt: 0.03 density ]{
\includegraphics[width=0.3\hsize]{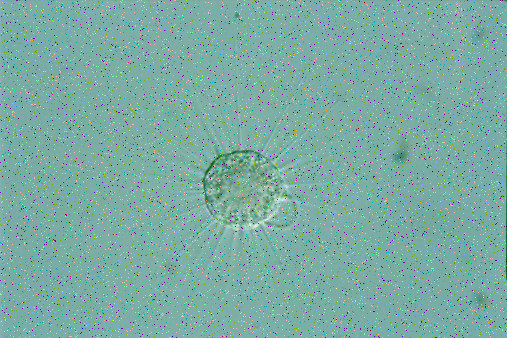}
}%
\centering
\caption{An example of different noisy EM images using EMDS-5 images. }
\label{fig:noise}
\end{figure}

We add a total of 13 kinds of noise (such as Gaussian noise and ``pepper and salt'' noise) 
to the original images and then denoise the noisy images with different methods. An 
example of the noisy EM images is shown in Fig.~\ref{fig:noise}.

We use nine different methods to denoise and choose to use the similarity 
between the denoised image and the original image and the variance of the two as the 
evaluation index. The evaluation index is expressed by Eq.~(\ref{eq:similarity})~\cite{Gonzalez-2007-Digital}.
\begin{align}
A=1-\frac{\sum_{1}^{n}\left|i_{1}-i\right|}{N\times255}.
\label{eq:similarity}
\end{align}

Where $A$ is the similarity, $i_{1}$ is the denoised image, $i$ is the original image, 
and $N$ is the number of pixels. The closer the value of $A$ is to 1, the better the 
denoising effect. We use the above original image as an example, and use the table to 
list the similarity between the image after removing various noises and the original 
image using different filters. The comparison of similarities between denoised images 
and original image using EMDS-5 are shown in TABLE~\ref{tab:noise-similarity}.
\begin{table*}[htbp!]
\scriptsize 
\centering
\caption{A comparison of similarities between denoised images and original image using EMDS-5. 
Types of noise (ToN), Denoising method (DM), Two-Dimensional Rank Order Filter (TROF), 
Mean Filter Window: $3 \times 3$ (MF: $3 \times 3$), 
Mean Filter Window: $5 \times 5$ (MF: $5 \times 5$), 
Wiener Filter Window: $3 \times 3$ (WF: $3 \times 3$), 
Wiener Filter Window: $5 \times 5$ (WF: $5 \times 5$), 
Maximum Filter (MaxF), Minimum Filter (MinF), Geometric Mean Filter (GMF), 
Arithmetic Mean Filter (AMF), Poisson noise (PN), 
Multiplicative noise variance: 0.2 (MN $v$: 0.2), 
Multiplicative noise variance: 0.04 (MN $v$: 0.04), 
Gaussian noise Variance: 0.01, Mean: 0 (GN $m$: 0, $v$: 0.01), 
Gaussian noise Variance: 0.01, Mean: 0.5 (GN m: 0.5, $v$: 0.01), 
Gaussian noise Variance: 0.03, Mean: 0 (GN $m$: 0, $v$: 0.03), 
Gaussian noise Variance: 0.03, Mean: 0.5 (GN m: 0.5, $v$: 0.03), 
Salt and pepper noise density: 0.01 (SPN $d$: 0.01), 
Salt and pepper noise density: 0.03 (SPN $d$: 0.03), 
Pepper noise (PpN), Brightness Gaussian noise (BGN), 
Position Gaussian noise (PGN), Salt noise (SN). (In [$\%$].)}
\begin{tabular}{| p{0.1\textwidth}<{\centering}| p{0.05\textwidth}<{\centering}| p{0.05\textwidth}<{\centering}| p{0.075\textwidth}<{\centering}| p{0.075\textwidth}<{\centering}| p{0.075\textwidth}<{\centering}| p{0.06\textwidth}<{\centering}| p{0.06\textwidth}<{\centering}| p{0.06\textwidth}<{\centering}| p{0.06\textwidth}<{\centering}|}
\hline
       ToN / DM & TROF & MF: $3 \times 3$ & MF: $5 \times 5$ & WF: $3 \times 3$ & WF: $5 \times 5$ & MaxF & MinF & GMF & AMF \\
\hline
      PN & 99.47 & 99.17 & 99.39 & 99.26 & 99.47 & 92.56 & 99.98 & 99.65 & 99.30\\
      \hline
      MN, $v$: 0.2 & 96.57& 94.34 & 96.43 & 96.56 & 98.11 & 68.55 & 99.90 & 98.61 & 97.08\\
      \hline
      MN, $v$: 0.04 & 98.27 & 97.14 & 98.16 & 97.91 & 98.73 & 81.13 & 99.94 & 98.66 & 98.21\\
      \hline
      GN, $m$: 0, $v$: 0.01 & 98.99 & 98.34 & 98.94 & 98.41 & 98.99 & 84.99 & 99.97 & 98.88 & 98.64\\
      \hline
      GN, $m$: 0.5, $v$: 0.01 & 61.51 & 61.13 & 60.96 & 62.05 & 62.05 & 60.18 & 64.50 & 62.10 & 62.35 \\
      \hline
      GN, $m$: 0, $v$: 0.03 & 98.33 & 97.21 & 98.21 & 97.32 & 98.36 & 75.24 & 99.95 & 98.54 & 97.81 \\
      \hline
      GN, $m$: 0.5, $v$: 0.03 & 62.00 & 61.56 & 61.21 & 63.95 & 63.96 & 60.18 & 68.78 & 64.22 & 64.23 \\
      \hline
      SPN, $d$: 0.01  & 99.77 & 99.79 & 99.69 & 99.60 & 99.57 & 97.00 & 99.98 & 99.53 & 99.59\\
      \hline
      SPN, $d$: 0.03 & 99.77 & 99.79 & 99.69 & 99.25 & 99.29 & 93.86 & 99.98 & 99.31 & 99.30\\ 
      \hline
      PpN & 99.78 & 99.80 & 99.70 & 99.81 & 99.80 & 98.77 & 99.98 & 99.70 & 99.80\\
      \hline
      BGN & 99.32 & 98.94 & 99.26 & 99.03 & 99.36 & 90.54 & 99.98 & 98.93 & 99.13\\
      \hline
      PGN & 99.05 & 98.44 & 98.98 & 98.33 & 98.81 & 85.86 & 99.97 & 99.16 & 98.70\\
      \hline
      SN & 99.79 & 99.81 & 99.71 & 99.82 & 99.84 & 98.77 & 99.98 & 98.77 & 99.82 \\        
\hline
\end{tabular}
\label{tab:noise-similarity}
\end{table*}

From the comparision in TABLE~\ref{tab:noise-similarity}, we find that EMDS-5 can support 
distinguishable evaluation for different denoising methods. For example, the maximum filtering 
effect is not very good, so it is not ideal for the denoising results of Gaussian noise and 
multiplicative noise, but it is still very good for the denoising results of salt and 
pepper noise and Poisson noise.

In addition, the mean variance of the denoised image and the original image is an indicator of stability of denoising mehods. The mean variance is expressed by 
Eq.~(\ref{eq:variance})~\cite{Gonzalez-2007-Digital}.
\begin{align}
S=\frac{\sum_{1}^{n}\left(l_{(i, j)}-B_{(i, j)}\right)^{2}}{\sum_{1}^{n} l_{(i, j)}^{2}}.
\label{eq:variance}
\end{align}
Where $l_{(i, j)}$ and $B_{(i, j)}$ represent the pixels corresponding to the original image 
after denoising, and $S$ represents the mean variance. The comparison of variances between 
denoised images and original image using EMDS-5 are shown in TABLE~\ref{eq:variance}.
\begin{table*}[htbp!]
\scriptsize 
\centering
\caption{A comparison of variances between denoised images and original image using EMDS-5. 
Types of noise (ToN), Denoising method (DM), Two-Dimensional Rank Order Filter (TROF), 
Mean Filter Window: $3 \times 3$ (MF: $3 \times 3$), 
Mean Filter Window: $5 \times 5$ (MF: $5 \times 5$), 
Wiener Filter Window: $3 \times 3$ (WF: $3 \times 3$), 
Wiener Filter Window: $5 \times 5$ (WF: $5 \times 5$), 
Maximum Filter (MaxF), Minimum Filter (MinF), Geometric Mean Filter (GMF), 
Arithmetic Mean Filter (AMF), Poisson noise (PN), 
Multiplicative noise variance: 0.2 (MN $v$: 0.2), 
Multiplicative noise variance: 0.04 (MN $v$: 0.04), 
Gaussian noise Variance: 0.01, Mean: 0 (GN $m$: 0, $v$: 0.01), 
Gaussian noise Variance: 0.01, Mean: 0.5 (GN $m$: 0.5, $v$: 0.01), 
Gaussian noise Variance: 0.03, Mean: 0 (GN $m$: 0, $v$: 0.03), 
Gaussian noise Variance: 0.03, Mean: 0.5 (GN $m$: 0.5, $v$: 0.03), 
Salt and pepper noise density: 0.01 (SPN $d$: 0.01), 
Salt and pepper noise density: 0.03 (SPN $d$: 0.03), 
Pepper noise (PpN), Brightness Gaussian noise (BGN), Position Gaussian noise (PGN), 
Salt noise (SN). (In [$\%$].)}
\begin{tabular}{| p{0.13\textwidth}<{\centering}| p{0.059\textwidth}<{\centering}| p{0.059\textwidth}<{\centering}| p{0.059\textwidth}<{\centering}| p{0.059\textwidth}<{\centering}| p{0.059\textwidth}<{\centering}| p{0.059\textwidth}<{\centering}| p{0.059\textwidth}<{\centering}| p{0.059\textwidth}<{\centering}| p{0.059\textwidth}<{\centering}|}
\hline
       ToN / DM & TROF & MF: $3 \times 3$ & MF: $5 \times 5$ & WF: $3 \times 3$ & WF: $5 \times 5$ & MaxF & MinF & GMF & AMF \\
\hline
       PN & 0.57 & 0.13 & 0.10 & 0.10 & 0.06 & 1.81 & 1.71 & 0.07 & 0.19 \\
      \hline
      MN, $v$: 0.2 & 5.13 & 5.54 & 2.53 & 2.68 & 1.14 & 27.70 & 25.82 & 3.69 & 2.17\\
      \hline
      MN, $v$: 0.04 & 1.64 & 1.41 & 0.67 & 0.75 & 0.34 & 10.39 & 7.21 & 0.68 & 0.67\\
      \hline
      GN, $m$: 0, $v$: 0.01 & 0.85 & 0.49 & 0.25 & 0.48 & 0.22 & 7.10 & 3.94 & 0.40 & 0.42\\
      \hline
      GN, $m$: 0.5, $v$: 0.01 & 42.16 & 42.41 & 42.85 & 40.15 & 40.13 & 44.96 & 35.96 & 40.02 & 39.78 \\
      \hline
      GN, $m$: 0, $v$: 0.03 & 1.60 & 1.41 & 0.64 & 1.39 & 0.60 & 18.72 & 9.81 & 1.87 & 0.99 \\
      \hline
      GN, $m$: 0.5, $v$: 0.03 & 41.04 & 41.47 & 42.29 & 36.38 & 36.15 & 44.97 & 28.71 & 35.67 & 35.92 \\
      \hline
      SPN, $d$: 0.01  & 0.49 & 0.02 & 0.05 & 0.59 & 0.39 & 2.20 & 1.11 & 4.53 & 0.20\\
      \hline
      SPN, $d$: 0.03 & 0.48 & 0.02 & 0.05 & 1.35 & 0.69 & 5.87 & 1.80 & 12.67 & 0.37\\
      \hline
      PpN & 0.63 & 0.03 & 0.05 & 1.42 & 0.71 & 0.16 & 2.39 & 16.78 & 0.39\\
      \hline
      BGN & 0.63 & 0.21 & 0.13 & 0.18 & 0.10 & 2.89 & 2.19 & 0.43 & 0.24\\
      \hline
      PGN & 0.86 & 0.49 & 0.25 & 0.67 & 0.39 & 7.02 & 3.96 & 0.20 & 0.42\\
      \hline
      SN & 0.49 & 0.03 & 0.06 & 1.84 & 0.84 & 0.16 & 3.74 & 0.18 & 0.54 \\
                    
\hline
\end{tabular}
\label{tab:noise-variance}
\end{table*}

From the comparison in TABLE~\ref{tab:noise-variance}, we find that our EMDS-5 is useful 
to test and evaluate image decisioning methods effectively. For example, increasing the 
mean value of Gaussian noise will result in greater variance between the denoised images 
and the original images, indicating that the results after denoising are not very stable.

\subsection{Evaluation of Edge Detection Methods}
\begin{figure}[htbp!]
\centering
\subfigure[Orginal image]{
\includegraphics[width=0.3\hsize]{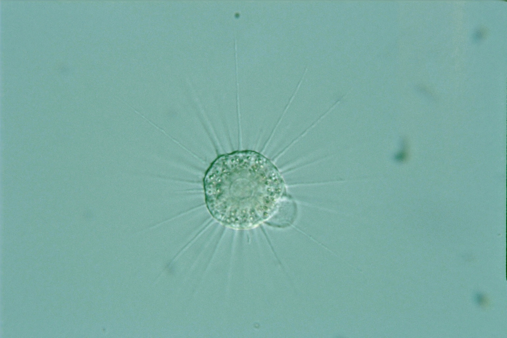}
}%
\quad
\subfigure[Edges in a GT image]{
\includegraphics[width=0.3\hsize]{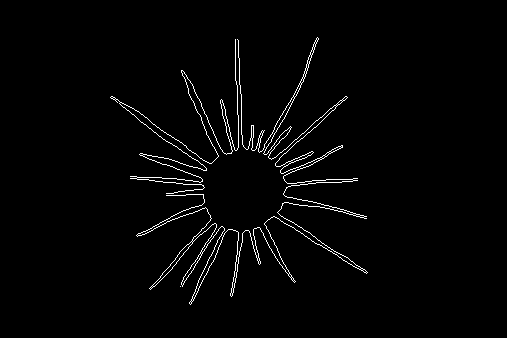}
}%
\quad
\subfigure[Sobel]{
\includegraphics[width=0.3\hsize]{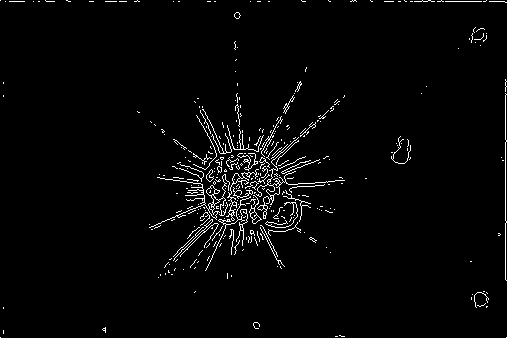}
}%
\quad
\subfigure[LoG]{
\includegraphics[width=0.3\hsize]{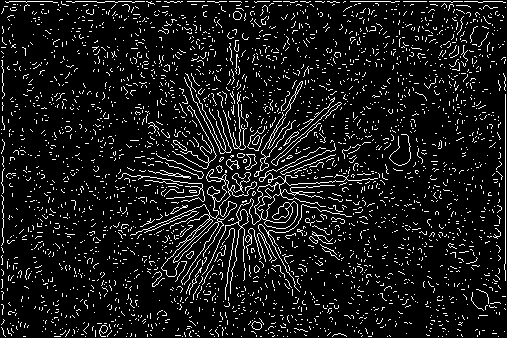}
}%
\quad               
\subfigure[Canny]{
\includegraphics[width=0.3\hsize]{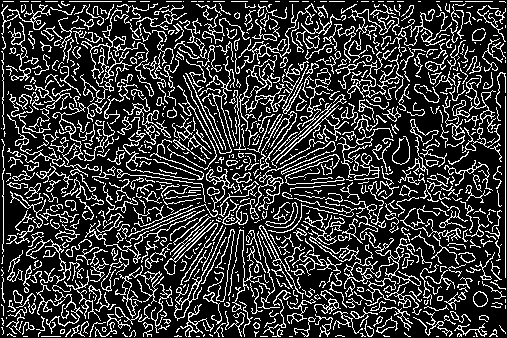}
}%
\quad 
\subfigure[Prewitt]{
\includegraphics[width=0.3\hsize]{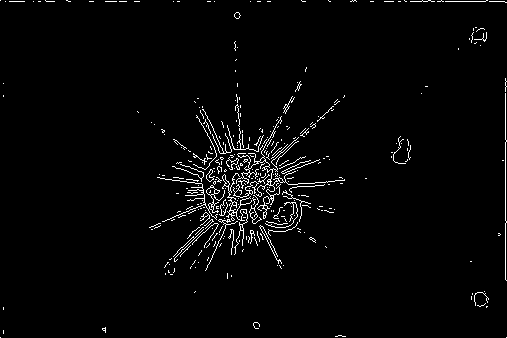}
}%
\quad 
\subfigure[Roberts]{
\includegraphics[width=0.3\hsize]{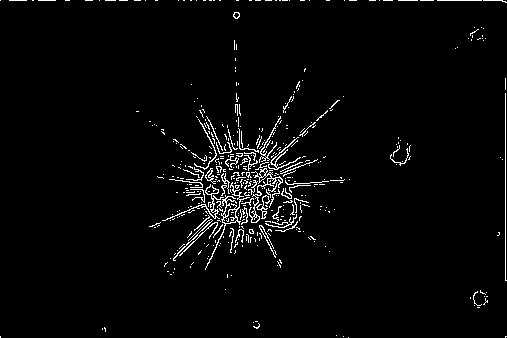}
}%
\quad 
\subfigure[Zero cross]{
\includegraphics[width=0.3\hsize]{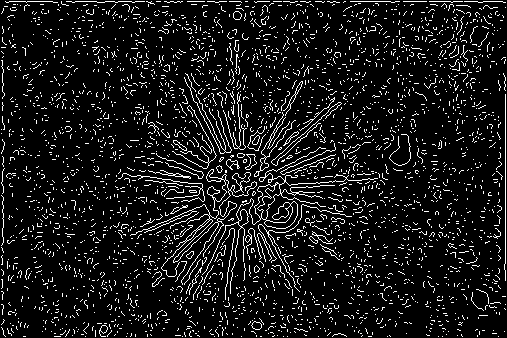}
}%
\centering
\caption{An example of six edge detection results using EMDS-5 images.}
\label{fig:edge}
\end{figure}

Edge detection is an important component of image preprocessing. In order to prove the 
effectiveness of our EMDS-5 in edge detecition evaluation, six operators are used to detect 
edges from images in EMDS-5 dataset. The six operators are Canny, Laplace of Gaussian (LoG), 
Prewitt, Roberts, Sobel and Zero cross, and an example of the edge detection results is 
shown in Fig.~\ref{fig:edge}.

Furthermore, we conduct a quantitative evaluation to compare edge detection results, 
where the edge detection results by Sobel operator on GT images are used as reference 
to evaluate other operators. Here we use the two indicators, \emph{Peak-Signal to Noise Ratio} 
(PSNR) and \emph{Mean Structural Similarity} (SSIM), to evaluate the results of edge 
detection. PSNR calculates the differences between the pixels' gray values corresponding 
to the image to be evaluated and the reference image to measure the quality of the image 
to be evaluated from a statistical point of view. We hypothesis that the image to be 
evaluated is $F$, the reference image is $R$, and their sizes are $MN$. The calculation 
method for characterizing image quality using PSNR, which is expressed 
by Eq.~(\ref{eq:psnr})~\cite{gupta-2011-modified}.
\begin{align}
\mathrm{PSNR}=10 \lg \frac{255^{2}}{\frac{1}{\mathrm{MN}} \sum_{i=1}^{M} \sum_{j=1}^{N}|R(i, j)-F(i, j)|^{2}}.
\label{eq:psnr}
\end{align}

PSNR measures the image quality by calculating the global size of the pixel error 
between the image to be evaluated and the reference image. The larger the PSNR value, 
the less distortion between the image to be evaluated and the reference image, and 
the image quality is better. SSIM is a commonly used image quality evaluation method 
originally proposed in~\cite{wang-2004-image}. SSIM is composed of three contrast functions. 
The brightness contrast function is expressed by Eq.~(\ref{eq:brightness}).
\begin{align}
l(x, y)=\frac{2 u_{x} u_{y}+c_{1}}{u_{x}^{2}+u_{y}^{2}+c_{1}}.
\label{eq:brightness}
\end{align}

Contrast contrast function is expressed by Eq.~(\ref{eq:contrast-c}). 
\begin{align}
c(x, y)=\frac{2 \sigma_{x} \sigma_{y}+c_{2}}{\sigma_{x}^{2}+\sigma_{y}^{2}+c_{2}}.
\label{eq:contrast-c}
\end{align}

Structural contrast function is expressed by Eq.~(\ref{eq:structural-c}).
\begin{align}
s(x, y)=\frac{\sigma_{x y}+c_{3}}{\sigma_{x} \sigma_{y}+c_{3}}.
\label{eq:structural-c}
\end{align}

$sigma_{x y}$ is expressed by Eq.~(\ref{eq:sigma}).
\begin{align}
\sigma_{x y}=\frac{1}{N-1} \sum_{i=1}^{N}\left(x_{i}-\mu_{x}\right)\left(y_{i}-\mu_{y}\right).
\label{eq:sigma}
\end{align}

We combine the three functions and finally get the SSIM index function expressed 
by Eq.~(\ref{eq:SSIM}).
\begin{align}
\operatorname{SSIM}(x, y)=l(x, y)^{a} \cdot c(x, y)^{\beta} \cdot s(x, y)^{?}.
\label{eq:SSIM}
\end{align} 
Where $u_{x}$, $u_{y}$ are all pixels of the image block; 
$\sigma_{x}$ ,$\sigma_{y}$ are the standard deviation of the image pixel values; 
$\sigma_{x} \sigma_{y}$ is the covariance of $x$ and $y$; 
$C_{1}$, $C_{2}$, $C_{3}$ are constants, in order to avoid the system error caused 
when the denominator is 0. 
SSIM is a number between 0 and 1. 
The larger the SSIM, the smaller the difference between the two images. 
A comparison of edge detection methods using EMDS-5 is shown in TABLE~\ref{tab:edge}.
\begin{table}[htbp!]
\centering
\caption{A comparison of edge detection methods using EMDS-5. 
Evaluation index (EI), Operator type (OT).}
\begin{tabular}{| p{0.1\textwidth}<{\centering}| p{0.1\textwidth}<{\centering}| p{0.1\textwidth}<{\centering}| p{0.1\textwidth}<{\centering}| p{0.1\textwidth}<{\centering}| p{0.1\textwidth}<{\centering}|} 
\hline
       EI / OT &  Canny & LoG & Prewitt & Roberts & Zero cross  \\
\hline
      \textit{PSNR} & 54.84 & 58.16 & 72.44 & 63.37 & 58.16  \\
      \hline
      \textit{SSIM} & 98.89$\%$ & 99.67$\%$ & 99.99$\%$ & 99.94$\%$ & 99.67$\%$ \\

\hline
\end{tabular}
\label{tab:edge}
\end{table}

From the TABLE~\ref{tab:edge}, we find that the PSNR evaluation index that the 
edge detection results obtained by the Prewitt operator are the most similar to 
the Sobel results. The SSIM evaluation index shows that the difference between 
the results of other operators and the results of Sobel operator is also very small. 
By comparison, we can see that EMDS-5 images can be used to detect and evaluate 
various edge detection methods.

\section{Image Segmentation Evaluation Using EMDS-5}
\subsection{Single-object Image Segmentation}
\begin{figure}[htbp!]
\centering
\subfigure[Orginal image]{
\includegraphics[width=0.3\hsize]{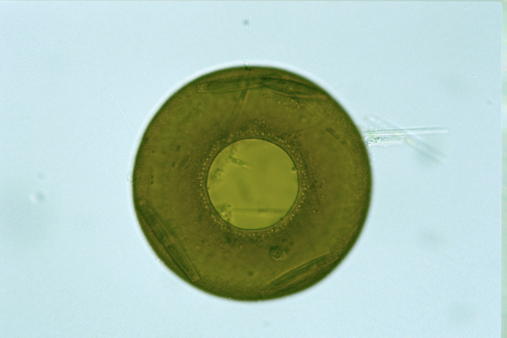}
}%
\quad
\subfigure[GT image]{
\includegraphics[width=0.3\hsize]{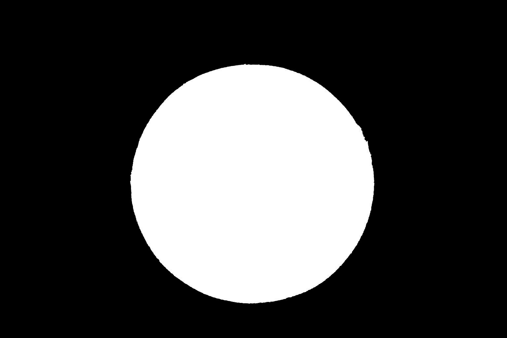}
}%
\quad
\subfigure[GrubCut]{
\includegraphics[width=0.3\hsize]{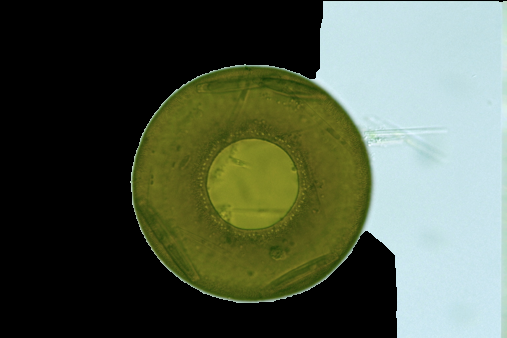}
}%
\quad
\subfigure[MRF]{
\includegraphics[width=0.3\hsize]{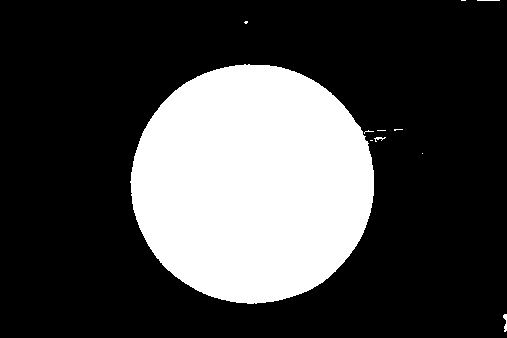}
}%
\quad               
\subfigure[Otsu thresholding]{
\includegraphics[width=0.3\hsize]{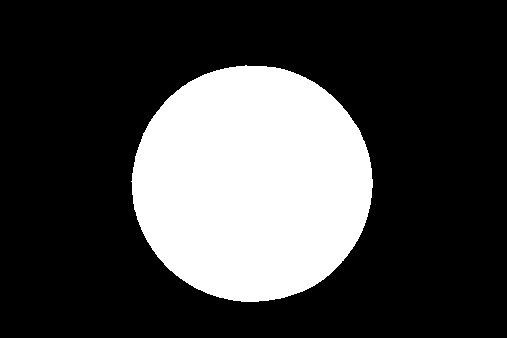}
}%
\quad 
\subfigure[Region growing]{
\includegraphics[width=0.3\hsize]{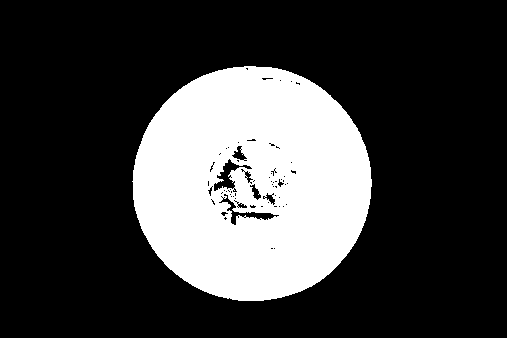}
}%
\quad
\subfigure[Canny edge detection based]{
\includegraphics[width=0.3\hsize]{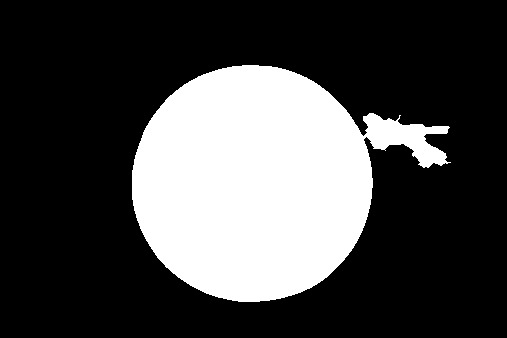}
}%
\quad
\subfigure[Watershed]{
\includegraphics[width=0.3\hsize]{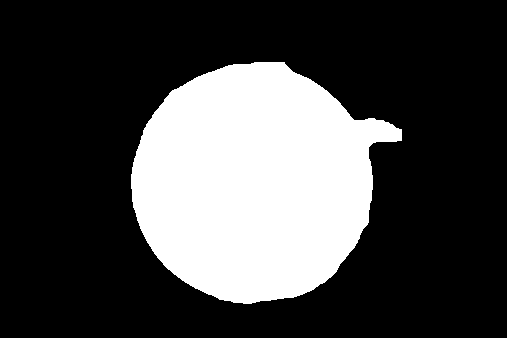}
}%
\quad
\centering
\caption{An example of different single-object segmentation results using EMDS-5 images.}
\label{fig:SO-example}
\end{figure}

In order to prove the effectiveness of EMDS-5 for image segmentation evaluation, 
six typical image segmentation methods are compared to segment the EMDS-5 original images, 
including GrubCut, Markov Random Field (MRF), Canny edge detection based, Watershed, 
Otsu thresholding and Region growing approaches. GrubCut is a common and classic method of 
semi-automatic segmentation. MRF is a classical graph based segmentation method.  
Otsu thresholding is an image segmentation method based on threshold. 
Region growing approaches and Watershed algorithm are classical region based segmentation methods. 
An example of different single-object segmentation results is shown in Fig.~\ref{fig:SO-example}.  

We compare the images obtained after image segmentation with the corresponding GT images, 
where five evaluation indexes in TABLE~\ref{tab:SO-seg} are used to evaluate the 
segmentation results~\cite{wang-2004-image,Taha-2015-Metrics}. 
\begin{table}[htbp!] 
\centering
 \caption{The definitions of evaluation metrics for image segmentation. 
 TP (True Positive), FN (False Negative), FP (False Positive).}
\setlength{\tabcolsep}{1cm}{ 
 \begin{tabular}{|c|c|}
  \hline
  Metric & Definition  \\ 
 \hline
 Dice & Dice $=\frac{2 \times\left|V_{\text {pred }} \cap V_{\mathrm{gt}}\right|}{\left|V_{\text {pred }}\right|+\left|V_{\mathrm{gt}}\right|}$ \\
 \hline
  Jaccard & Jaccard $=\frac{\left|V_{\text {pred }} \bigcap V_{\mathrm{gt}}\right|}{\left|V_{\text {pred }} \bigcup V_{\mathrm{gt}}\right|}$ \\ 
 \hline
  Recall & $\operatorname{Recall}=\frac{\mathrm{TP}}{\mathrm{TP}+\mathrm{FN}}$ \\ 
 \hline
 \end{tabular}}
 \label{tab:SO-seg}
\end{table}

In TABLE~\ref{tab:SO-seg}, $V_{\text {pred }}$ represents the foreground that is 
predicted by the model; $V_{\mathrm{gt}}$ represents the foreground in a ground truth image. 
We show the evaluation results of the sample images in Table~\ref{tab:SO-index}.
\begin{table}[htbp!]
\centering
\caption{A comparison of single-object segmentation methods using EMDS-5. 
Image segmentation methods(ISM), Evaluation index (EI), 
Watershed algorithm (WA), Otsu thresholding (OT), Region growing (RG). (In [$\%$].)}
\begin{tabular}{| p{0.1\textwidth}<{\centering}| p{0.1\textwidth}<{\centering}| p{0.1\textwidth}<{\centering}| p{0.1\textwidth}<{\centering}|} 
\hline
       ISM / EI &   Dice & Jaccard & Recall  \\
\hline
      GrubCut & 18.41 & 10.14 & 10.18 \\
      \hline
      MRF & 98.01 & 96.09 & 99.67  \\
      \hline
      Canny & 59.59 & 51.48 & 94.99  \\
      \hline
      WA & 57.79 & 49.50 & 76.26 \\
      \hline
      OT & 98.87 & 97.76 & 98.16 \\
      \hline
      RG & 86.67 & 76.47  & 77.65  \\           
      \hline
\end{tabular}
\label{tab:SO-index}
\end{table}

From TABLE~\ref{tab:SO-index}, it is observed that because the GrubCut method segments 
the original image, the result obtained when compared with the GT image will lead to a 
low evaluation result. Among several other classic single-object image segmentation methods, 
the results of Otsu Thresholding and MRF segmentation closest to the GT image and the 
best effect. Other segmentation methods have a certain gap compared with these two 
segmentation methods. Through the comparison of these image segmentation parties, 
we can conclude that EMDS-5 is effective in testing and evaluating image segmentation 
methods.

\subsection{Multi Object Image Segmentation}
\begin{figure}[htbp!]
\centering
\subfigure[Orginal Image]{
\includegraphics[width=0.3\hsize]{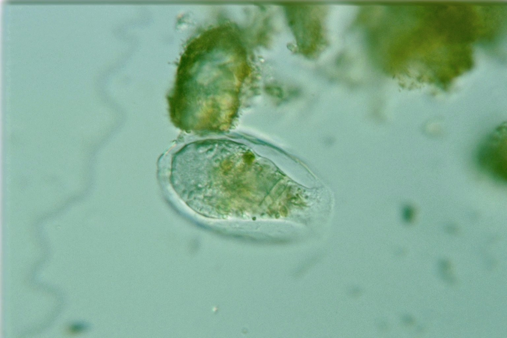}
}%
\quad
\subfigure[GT Image]{
\includegraphics[width=0.3\hsize]{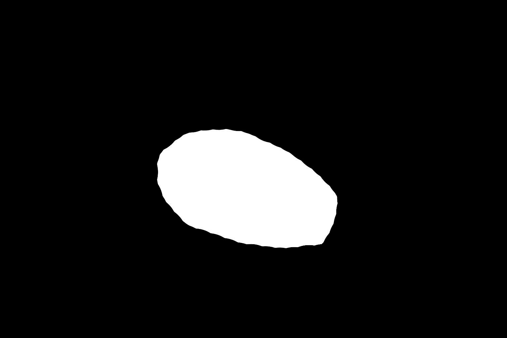}
}%
\quad
\subfigure[ $\mathit{k}$-means]{
\includegraphics[width=0.3\hsize]{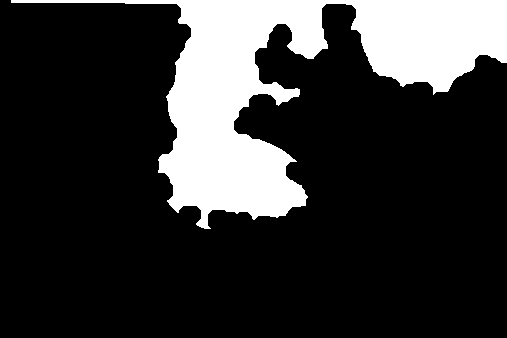}
}%
\quad
\subfigure[U-net]{
\includegraphics[width=0.3\hsize]{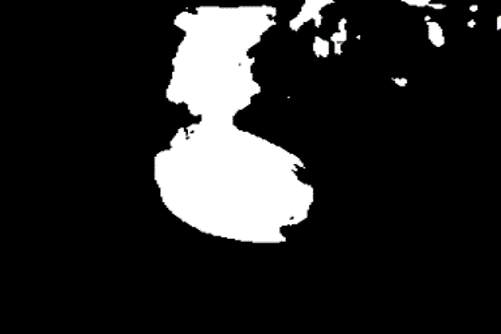}
}%
\quad
\centering
\caption{An example of different multi-object segmentation results using EMDS-5. }
\label{fig:MO-seg}
\end{figure}

For multi-object image segmentation, we use two methods, $k$-means and U-net, to test 
our EMDS-5. $k$-means is an unsupervised learning approach (clustering) and U-net is 
a supervised learning method (deep convolutional neural Network). These two methods 
are representative of the classic methods in their respective fields. The examples of 
different multi object segmentation methods results are shown in Fig.~\ref{fig:MO-seg}.

For these two multi-object image segmentation methods, a comparison is shown in 
TABLE~\ref{tab:MO-index} .
\begin{table}[htbp!]
\centering
\caption{A comparison of multi-object segmentation methods using EMDS-5. 
Image segmentation methods (ISM), Evaluation index (EI). (In [$\%$].)}
\begin{tabular}{| p{0.1\textwidth}<{\centering}| p{0.1\textwidth}<{\centering}| p{0.1\textwidth}<{\centering}| p{0.1\textwidth}<{\centering}|} 
\hline
       ISM / EI & Dice & Jaccard & Recall  \\
\hline
      $k$-means & 31.97 & 25.93 & 65.81 \\
      \hline
      U-net & 85.24 & 77.41 & 82.28  \\         
      \hline
\end{tabular}
\label{tab:MO-index}
\end{table}

It can be seen from TABLE~\ref{tab:MO-index} that the segmentation effect of U-net 
in the multi-target image segmentation method is much higher than that of K-meansm, 
showing the effectiveness of EMDS-5 for evaluaiton of multi-object image segmentation 
methods.

\section{Feature Extraction Evaluation Using EMDS-5}
\label{sec:feature}
We use GT images to localize the target EMs in the original images to test feature 
extraction methods. Since GT images have single-object GT images and multi-object GT 
images, feature extraction methods are grouped into two types. An example of original 
images and target EM images extracted from GT images are shown in Fig. 7.
\begin{figure}[htbp!]
\centering
\subfigure[An original image with a single EM]{
\includegraphics[width=0.3\hsize]{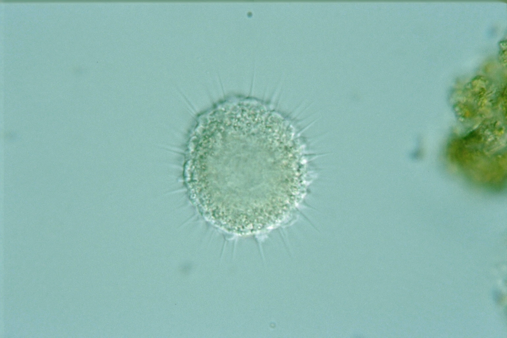}
}%
\quad
\subfigure[Localized single EM]{
\includegraphics[width=0.3\hsize]{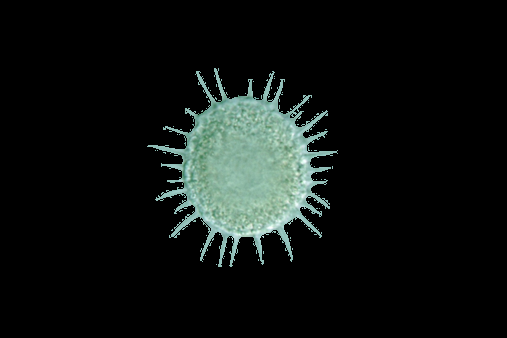}
}%
\quad               
\subfigure[An original image with multiple EMs]{
\includegraphics[width=0.3\hsize]{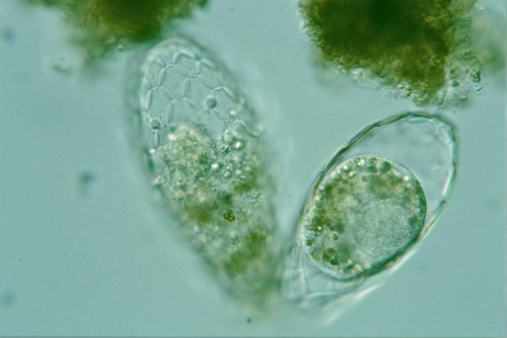}
}%
\quad 
\subfigure[Localized multiple EMs]{
\includegraphics[width=0.3\hsize]{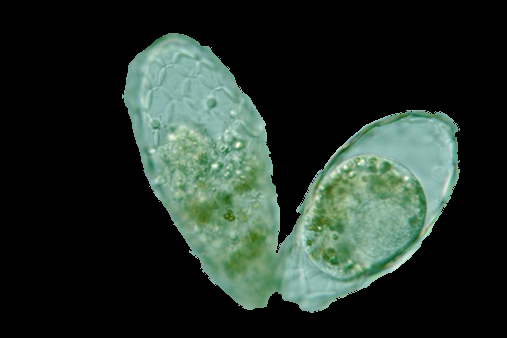}
}%
\centering
\caption{An example of localized EMs by GT images.}
\end{figure}

First, we randomly select ten images from each EM class as the training set and 
the other ten as the test set. Then, we extract and compare 12 features, including 
two color feaures (HSV (Hue, Saturation and Value) and RGB (Red, Green and Blue) features), 
three texture features (GLCM (grey-level co-occurrence matrix), HOG (Histogram of 
Oriented Gridients) and LBP (Local Binary Pattern) features), 
four geometric features (area, perimeter, long and short axis features), 
seven invariant moment features (Hu moments), 
and two deep learning features (VGG16 and Resnet50 features). 
We test the color features extract from the respective channels of RGB features and 
HSV features as a single feature vector. 
Lastly, we use a Radial Basis Function Support Vector Machine (RBFSVM) 
classifier (supported by LIBSVM~\cite{chang-2011-libsvm}) to test each feature and 
calculate their accuracies. 
The LIBSVM parameters are set as $-s$ 0 $-t$ 0 $-c$ 2 $-g$ 1 $-b$ 1. 
%$-s$ SVM type : set type of SVM (default 0). 
%$-t$ kernel type : set type of kernel function (default 2). 
%$-c$ cost : set the parameter C of C-SVC, epsilon-SVR, and nu-SVR (default 1). 
%$-g$  gamma: set the parameters of the polynomial, RBF, and sigmoid kernel functions. 
%$-b$ probability estimates: whether to train a SVC or SVR model for probability estimates, 0 or 1 (default 0).

\subsection{Single-Object Feature Extraction}
In TABLE~\ref{tab:SO-feature}, the accuracies of EM image classification using 
single-object features are compared. 
\begin{table}[htbp!]
\centering
\caption{Classification accuracy of single-object features by RBFSVM using EMDS-5.F
Feature type (FT), Accuracy (Acc), Geometric features (Geo), Hu moments (Hu). (In [$\%$].)}
\begin{tabular}{| p{0.043\textwidth}<{\centering}
| p{0.1\textwidth}<{\centering}
| p{0.1\textwidth}<{\centering}
| p{0.1\textwidth}<{\centering}
| p{0.1\textwidth}<{\centering}
| p{0.1\textwidth}<{\centering}
| p{0.1\textwidth}<{\centering}|} 
\hline
       FT  & RGB-R & RGB-G & RGB-B & HSV-H & HSV-S & HSV-V  \\
\hline
      \textit{Acc} & 27.62 & 36.67 & 34.76 & 30.48 & 34.29 & 39.52 \\
      \hline
       Geo & Hu & LBP & HOG & GLCM & VGG16 & Resnet50  \\
      \hline
       41.43 & 7.62 & 38.01 & 10 & 28.10 & 83.81 & 39.45\\
      \hline
\end{tabular}
\label{tab:SO-feature}
\end{table}

\subsection{Multi-object Feature Extraction}
In TABLE~\ref{tab:MO-feature}, the accuracies of EM image classification using 
multi-object features are compared. 
\begin{table}[htbp!]
\centering
\caption{Classification accuracy of multi-object features by RBFSVM using EMDS-5.F
Feature type (FT), Accuracy (Acc), Geometric features (Geo), Hu moments (Hu). (In [$\%$].)}
\begin{tabular}{| p{0.043\textwidth}<{\centering}
| p{0.1\textwidth}<{\centering}
| p{0.1\textwidth}<{\centering}
| p{0.1\textwidth}<{\centering}
| p{0.1\textwidth}<{\centering}
| p{0.1\textwidth}<{\centering}
| p{0.1\textwidth}<{\centering}|} 
\hline
       FT  & RGB-R & RGB-G & RGB-B & HSV-H & HSV-S & HSV-V  \\
\hline
      \textit{Acc} & 22.86 & 29.05 & 28.57 & 28.57 & 29.05 & 33.81 \\
      \hline
      Geo & Hu & LBP & HOG & GLCM & VGG16 & Resnet50  \\
      \hline
      38.10 & 7.62 & 37.62 & 14.76 & 22.38 & 68.57 & 23.33 \\
      \hline
\end{tabular}
\label{tab:MO-feature}
\end{table}

From TABLE~\ref{tab:SO-feature} and~\ref{tab:MO-feature}, we can find that when using 
the same RBFSM classifiers to classify EM images with different features, we obtain 
different classification results, showing the effectiveness of EMDS-5 for the feature 
extraction evaluation. Especially, because VGG16 feature achieves the best effect, 
we chose it in the following section about ``classification evaluation''.

\section{Image Classification Evaluation Using EMDS-5}
We use the features extracted from the EMDS-5 data to test classification performance 
of different classifiers. As mentioned in Sec.~\ref{sec:feature}, we use the extracted 
VGG16 features for testing in this section. The VGG16 feature vector selects the 16th 
layer feature vector. The dimension is 1$\times$1000. 
First, we randomly select ten images from each EM class as the training set and use 
another ten as test set. 
Then, we select 14 normally used classifiers for EM image classification, including 
four SVMs, three $k$-Nearest Neighbors (KNNs), three Random Forests (RFs), 
two VGG16 and two Inception-V3 classifiers. 
We combine and compare the extracted VGG16 features with four classic classifiers. 
In addition, four deep learning classifiers are directly compared. In VGG16 and Inception-V3, 
we divide the data into test, validation and test sets. Then we test the accuracy of 
any two types of EM image classification. We change the ratio of the images owned by 
these three datasets and test the accuracy, separately. 
Especially, the parameters of four SVM classifiers are shown in TABLE~\ref{tab:classifier}.
 \begin{table}[htbp!] 
 \centering
 \caption{The parameters of four SVM classifiers for EMDS-5 image classification 
 (supported by LIBSVM). }
\setlength{\tabcolsep}{1cm}{ 
 \begin{tabular}{|c|c|} 
  \hline 
 SVM type & Parameter  \\ 
  \hline
 SVM: linear & $-s$ 0 $-t$ 0 $-c$ 2 $-g$ 1 $-b$ 1 \\
 \hline
 SVM: polynomial & $-s$ 0 $-t$ 1 $-r$ 0 $-g$ 0.42 $-d$ 3\\
 \hline
  SVM: RBF & $-s$ 0 $-t$ 2 $-c$ 2 $-g$ 1 $-b$ 1 \\
  \hline
   SVM: sigmoid & $-s$ 0 $-t$ 3 $-r$ 0 $-g$ 0.042 \\ 
 \hline
 \end{tabular}}
 \label{tab:classifier}
\end{table}

Furthermore, a comparison of different classifiers for EM image classification 
using EMDS-5 is shown in TABLE~\ref{tab:class-acc}. 
\begin{table}[htbp!] \small
\centering
\caption{ A comparison of EM image classification results using EMDS-5. 
Accuracy (Acc), $n$Tree ($n$T), 
VGG16 (Train : Validation : Test = 1 : 1 : 2) is VGG16: 1 : 1 : 2, 
VGG16 (Train : Validation : Test = 1 : 2 : 1) is VGG16: 1 : 2 : 1, 
Inception-V3 (Train : Validation : Test = 1 : 1 : 2) is I-V3: 1 : 1 : 2, 
Inception-V3 (Train : Validation : Test = 1 : 2 : 1) is I-V3: 1 : 2 : 1. (In [$\%$].)}
  \begin{tabular} 
{| p{0.1\textwidth}<{\centering}
| p{0.1\textwidth}<{\centering}
| p{0.1\textwidth}<{\centering}
| p{0.1\textwidth}<{\centering}
| p{0.1\textwidth}<{\centering}|}
\hline
       Classifier type  & SVM: linear & SVM: polynomial & SVM: RBF & SVM: sigmoid   \\
\hline
      Acc & 68.57 & 63.81 & 21.91 & 5.24  \\
      \hline    
      $k$-NN, $k$: 1 &  $k$-NN, $k$: 5 & $k$-NN, $k$: 10 &  RF, $n$T: 10 & RF, $n$T: 20  \\
      \hline
      60.48 & 52.38 & 48.10 & 44.76 & 47.14  \\
      \hline
     RF, $n$T: 30 & VGG16, 1:1:2 & VGG16, 1:2:1 & I-V3, 1:1:2 & I-V3, 1:2:1 \\
      \hline
     55.71 & 81.61 & 83.23 & 89.43 & 90.49 \\
      \hline   
\end{tabular}
\label{tab:class-acc}
\end{table}

It can be seen from  the TABLE~\ref{tab:class-acc} that when using the same feature 
to test different classifiers, the deep learning network works best. The classification 
results of the two deep learning networks are the best. From the comparison of the results 
of different classifiers, we can see that EMDS-5 images can be effectively applied to 
the testing and evaluation of various classification algorithms.

\section{Image Retrieval Evaluation Using EMDS-5}
We use EMDS-5 for image retrieval. Because we use different features, 
we group the image retrieval methods into two categories: texture feature 
and deep learning feature based image retrieval approaches. We use 
Average Precision (AP)~\cite{Zou-2016-Content} to evaluate the retrieval 
results. $AP$ is developed in the field of information retrieval and is 
used to evaluate a ranked list of retrieved samples. The definition of AP 
in our article is shown in Eq.~(\ref{eq:AP}).
\begin{align}
\mathrm{AP}=\frac{\sum_{i=1}^{n}(P(k) \times \operatorname{rel}(k))}{M}.
\label{eq:AP}
\end{align}

$M$ is the number of related EM images, $P(k)$ is by considering the cut-off 
position divided by the $k$th position in the list, and $\operatorname{rel}(k)$ 
is an index. The EM image rank in the $k$th position is the target type image, 
then take 1; otherwise, take 0. AP represents the average value of the accuracy 
of the current position target type EM image. Our experiment is conducted on 21 
types of EM images, so we apply the \emph{mean AP} (mAP) to summarize the APs 
of each class. It is calculated by obtaining the average value of AP. During the 
retrieval process, we match the feature vector of the image to be tested with 
the feature vectors of all the images in the EMDS-5 dataset and calculate the 
Euclidean distance between the two. Then calculate the mAP value of the search 
result of the type of image to be tested as the search result. We display the 
first 20 images in the search results, in which the frame of the correct image 
is marked with a color.

\subsection{Texture Feature based Image Retrieval Using EMDS-5}
\begin{figure}[htbp!]
\centering
\subfigure{
\includegraphics[width=0.9\hsize]{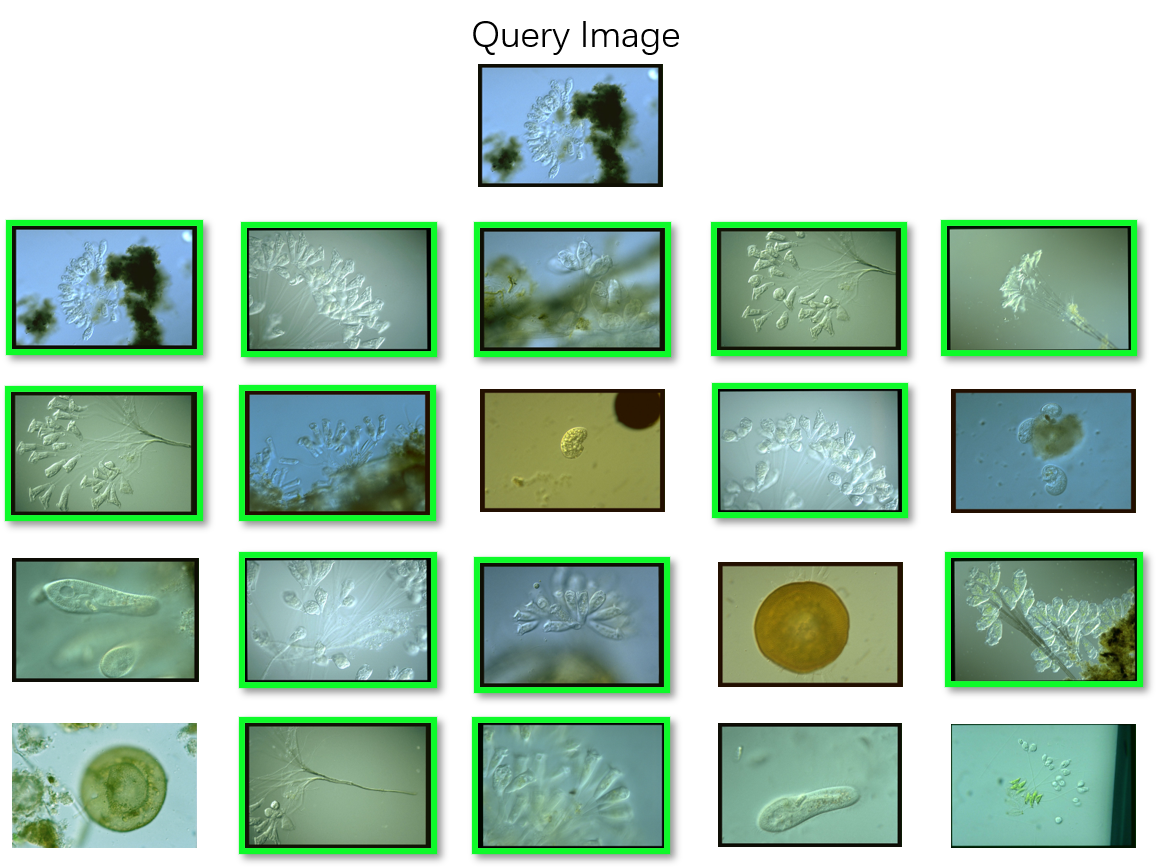}
}%
\centering
\caption{An example of image retrieval results with four texture features 
using EMDS-5.}
\label{fig:texture-retrieval}
\end{figure}

We extract a total of four texture features, GLCM, GGCM, HOG and LBP to test 
the EMDS-5 image retrieval evaluation function. An example of the image retrieval 
results based on texture features is shown in Fig.~\ref{fig:texture-retrieval}.

\begin{figure}[htbp!]
\centering
\subfigure[GLCM]{
\includegraphics[width=0.43\hsize]{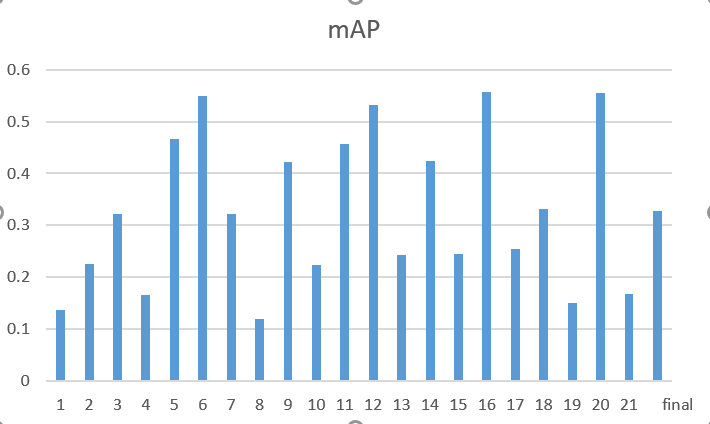}
}%
\quad
\subfigure[GGCM]{
\includegraphics[width=0.43\hsize]{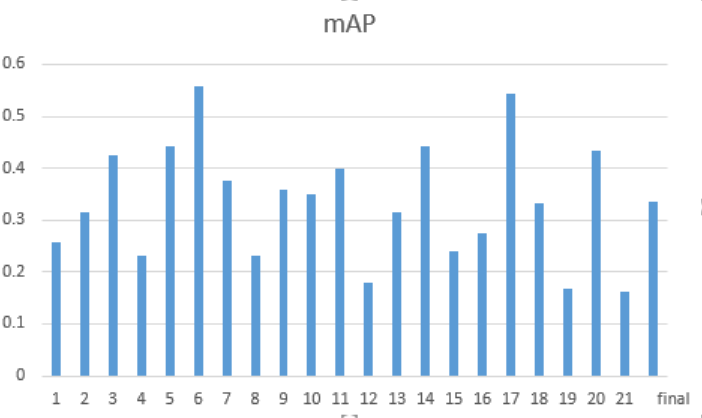}
}%
\quad
\subfigure[HOG]{
\includegraphics[width=0.43\hsize]{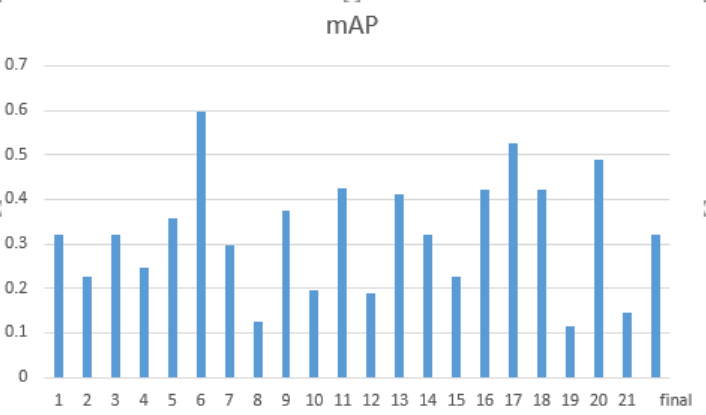}
}%
\quad
\subfigure[LBP]{
\includegraphics[width=0.43\hsize]{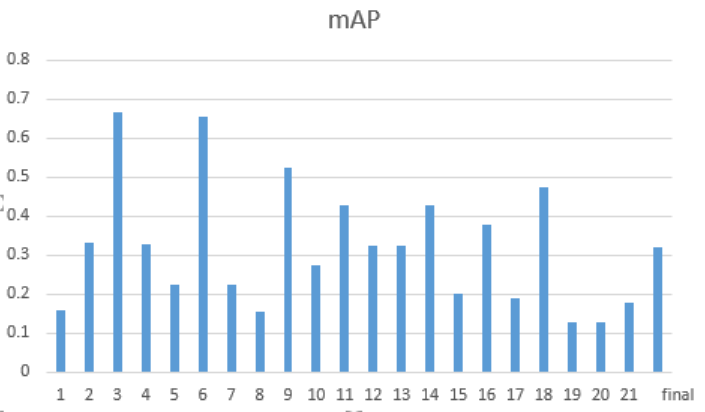}
}%
\quad
\centering
\caption{A comparison of image retrieval results with four texture features 
using EMDS-5.}
\label{fig:texture-retrieval-hist}
\end{figure}

Furthermore, the retrieval results of four texture features are demonstrated 
in Fig.~\ref{fig:texture-retrieval-hist}.

\subsection{Deep Learning Feature based Image Retrieval Using EMDS-5}
\begin{figure}[htbp!]
\centering
\subfigure{
\includegraphics[width=0.9\hsize]{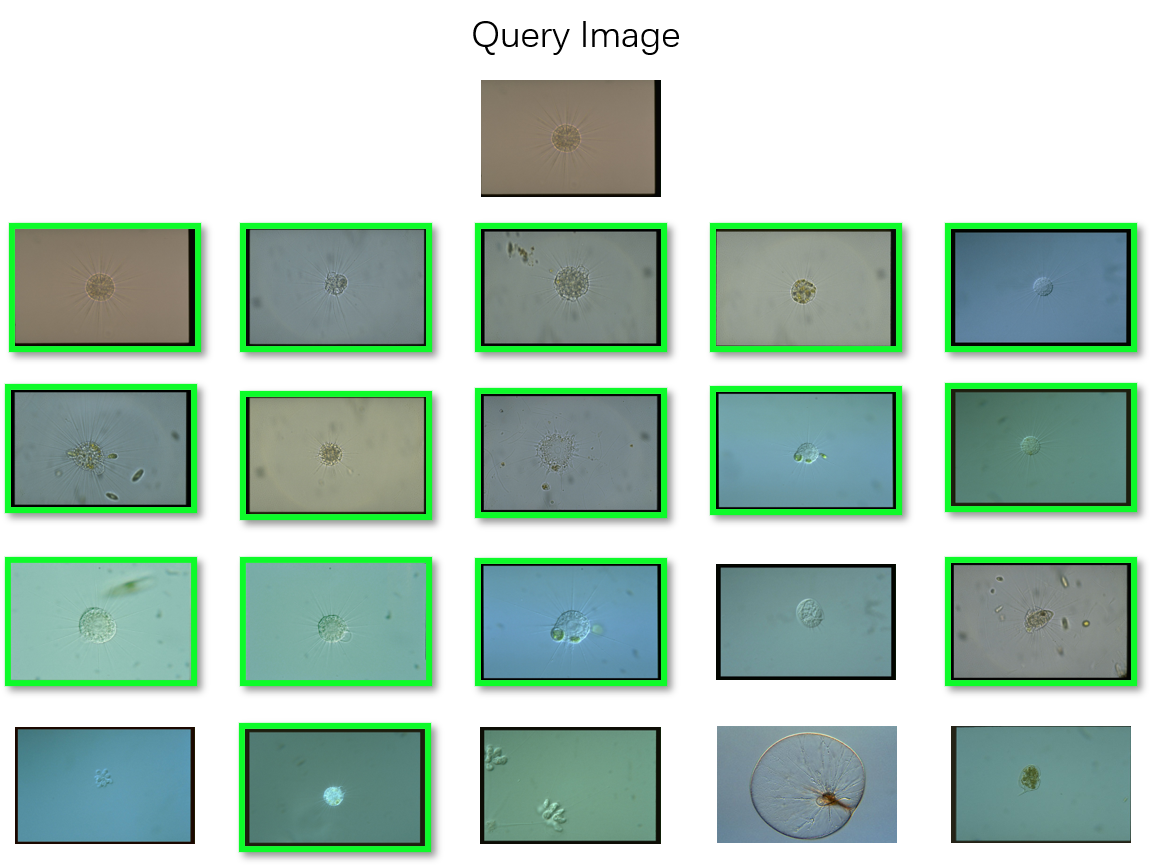}
}%
\centering
\caption{An example of image retrieval results based on deep learning features 
using EMDS-5.}
\label{fig:dl-retrieval}
\end{figure}

We first extract VGG16 features and Resnet50 features. Then, the selected feature 
vectors are the feature vectors of the last layer of the respective network. The 
dimension is $1 \times 1000$. The following figure is an example of retrieval results 
based on deep learning features. An example of retrieval results based on deep 
learning features is shown in Fig.~\ref{fig:dl-retrieval}.

\begin{figure}[htbp!]
\centering
\subfigure[VGG16]{
\includegraphics[width=0.43\hsize]{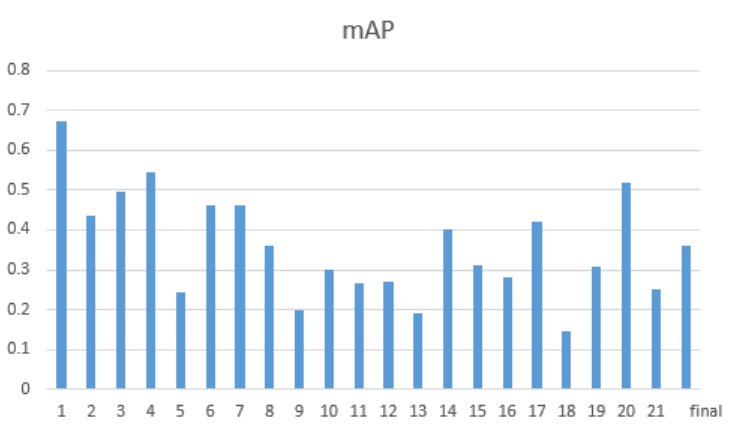}
}%
\quad
\subfigure[Resnet50]{
\includegraphics[width=0.43\hsize]{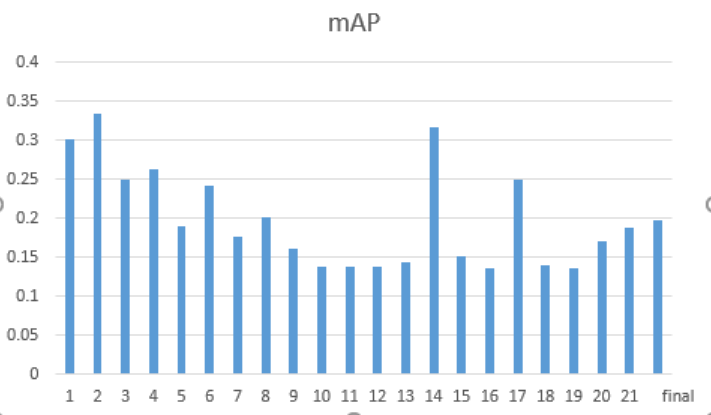}
}%
\quad
\centering
\caption{A comparison of image retrieval results with two deep learning features 
using EMDS-5.}
\label{fig:dl-retrieval-hist}
\end{figure}

Furthermore, the image retrieval results with two deep learning features are 
shown in Fig.~\ref{fig:dl-retrieval-hist}.

We calculate the variance of the mAP based on texture feature image retrieval 
and the variance of the $mAP$ based on deep learning feature image retrieval. 
The result we get is that the variance of the image retrieval results based on 
deep learning features is smaller, which shows that the results of deep learning 
feature image retrieval are more stable. By comparing the results of different 
retrieval methods, we can know that EMDS-5 images can be effectively applied to 
various image retrieval tests and evaluations.
\section{Conclusion and Future Work}
EMDS-5 is a microorganism dataset containing 21 microorganisms. EMDS-5 contains 
the original image and GT images of each EM. GT images include 
single-object GT images and multi-object GT images. Each original image has two 
corresponding GT images. Each microorganism has 20 original images, 20 single-object 
GT images and 20 multi-object GT images. EMDS-5 has the function of testing the 
denoising effect. When testing the denoising effect of EMDS-5, we add 13 kinds 
of noise, such as Possion noise and Gaussian noise. We use nine kinds of filters 
to test the denoising effect of various noises and achieved good results. EMDS-5 
can also evaluate and test the results of edge detection methods. We adopt six 
edge detection methods and use two evaluation indexes to evaluate the detection 
results and get good results. In terms of image segmentation, EMDS-5 can detect 
the results of image segmentation due to its single-object GT image and multi-object 
GT image. So we do  the testing with two parts: single-object image segmentation 
and multi-object image segmentation. In the single-object image segmentation part, 
we use six methods such as GrubCut and MRF to segment the original images and get 
good results. In terms of multi-object image segmentation, we use $k$-means and 
U-net methods for segmentation. We extract nine features from the images in the 
EMDS-5 database, such as RGB, HSV, GLCM, HOG. We use the LIBSVM classifier to detect 
the results of the extracted features. In the test, we randomly select ten images 
of each type of EMs as the training set and ten images as the test set. 
In terms of classification, we use the best VGG16 features to test different 
classifiers such as LIBSVM, KNN, RF. In terms of image retrieval, we divide image 
retrieval based on texture features and image retrieval based on deep learning features. 
In terms of texture features, we select four features, GLCM, GGCM, HOG and LBP, 
to test separately. In the deep learning feature, we select two deep learning features, 
VGG16 feature and Resnet50, for retrieval. We select the last layer of features of 
these two deep learning networks as feature vectors. We use mAP as an evaluation 
index to detect the quality of retrieval.

In the future, we will expand the types of microorganisms and increase the number 
of images of each microorganism. We hope that we can use the EMDS database to achieve 
more functions in the future.
\section*{Acknowledgments}
This work was supported in part by the National
Natural Science Foundation of China under Grant 61806047, in part
by the Fundamental Research Funds for the Central Universities under
Grant N2019003, and in part by the China Scholarship Council under
Grant 2017GXZ026396 and Grant 2018GBJ001757. We thank Prof.
Dr. Beihai Zhou and Dr. Fangshu Ma from the University of Science
and Technology Beijing, PR China, Prof. Joanna Czajkowska from
Silesian University of Technology, Poland, and Prof. Yanling Zou from
Freiburg University, Germany, for their previous cooperations in this
work.We also thank Miss Zixian Li and Mr. Guoxian Li for their important
discussion.

\nolinenumbers

% Either type in your references using
% \begin{thebibliography}{}
% Text
% \end{thebibliography}
%
% or
%
% Compile your BiBTeX database using our plos2015.bst
% style file and paste the contents of your .bbl file
% here. See http://journals.plos.org/plosone/s/latex for 
\bibliography{lizihan}

\end{document}